%% file: main.tex
\documentclass[10pt,twocolumn,letterpaper]{article}

\usepackage{iccv}              
\input{preamble}

\definecolor{iccvblue}{rgb}{0.21,0.49,0.74}
\usepackage[pagebackref,breaklinks,colorlinks,allcolors=iccvblue]{hyperref}

\def\paperID{897} 
\def\confName{ICCV}
\def\confYear{2025}

\title{LEGION: Learning to Ground and Explain for Synthetic Image Detection}

\author{
     Hengrui Kang$^{1,2}$\footnotemark[1], Siwei Wen$^{3,2}$\footnotemark[1], Zichen Wen$^{1,2}$\footnotemark[1], Junyan Ye$^{4,2}$, Weijia Li$^{4,2}$\footnotemark[2] \\  Peilin Feng$^{3,2}$, Baichuan Zhou$^{2}$, Bin Wang$^{2}$, Dahua Lin$^{2,5}$, Linfeng Zhang$^{1}$, Conghui He$^{2,5}$\footnotemark[2] \vspace{0.3em} \\
     {\normalsize $^1$Shanghai Jiao Tong University \quad $^2$Shanghai Artificial Intelligence Laboratory} \\
     {\normalsize $^3$Beihang University \quad $^4$Sun Yat-Sen University \quad $^5$SenseTime Research} \\
    {\normalsize Project Page: \url{https://opendatalab.github.io/LEGION}} \\
    {\normalsize E-mail: \texttt{liweij29@mail.sysu.edu.cn, heconghui@pjlab.org.cn}}
}

\begin{document}

\maketitle
{
\renewcommand{\thefootnote}{\fnsymbol{footnote}}
\footnotetext[1]{Equal Contribution.}
}
{
\renewcommand{\thefootnote}{\fnsymbol{footnote}}
\footnotetext[2]{Corresponding authors.}
}
\input{sec/0_abstract}

\input{sec/1_introduction}
\input{sec/2_relatework}
\input{sec/3_datasets}

\input{sec/4_method}
\input{sec/5_experiments}

\input{sec/6_conclusion}
\input{sec/appendix}

\clearpage
{
    \small
    \bibliographystyle{ieeenat_fullname}
    \bibliography{main}
}

\clearpage

\end{document}

%% file: preamble.tex
\newcommand{\VarSty}[1]{\textnormal{\ttfamily\color{blue!90!black}#1}\unskip}

\usepackage{bbding}

\usepackage{multirow}
\usepackage{makecell}
\usepackage{xcolor}
\usepackage{colortbl}

\usepackage{pifont}
\usepackage{amssymb}
\usepackage{fontawesome}
\usepackage{bm}
\usepackage{tcolorbox}

\definecolor{tkcolor}{RGB}{224,223,255}
\definecolor{tkcolor2_back}{RGB}{242,242,242}
\definecolor{tkcolor2_frame}{RGB}{50,50,50}
\definecolor{darkgreen}{RGB}{0,200,0}
\newtcolorbox{takeaways}[2][]{
	width=\columnwidth,
        toprule=0.0pt,
        leftrule=0.9pt,
        bottomrule=0.9pt,
        rightrule=0.9pt,
        arc=0pt,
	colback = tkcolor2_back, 
	colframe = tkcolor2_frame, 
	boxsep=0pt,left=7pt,right=7pt,top=4pt,bottom=4pt,
	fontupper=\linespread{1.1}\selectfont,
	title=#2,#1}

%% file: sec/0_abstract.tex
\begin{abstract}
The rapid advancements in generative technology have emerged as a double-edged sword. While offering powerful tools that enhance convenience, they also pose significant social concerns. 
As \textbf{defenders}, current synthetic image detection methods often lack artifact-level textual interpretability and are overly focused on image manipulation detection, and current datasets usually suffer from outdated generators and a lack of fine-grained annotations. 
In this paper, we introduce \textbf{SynthScars}, a high-quality and diverse dataset consisting of 12,236 fully synthetic images with human-expert annotations. It features 4 distinct image content types, 3 categories of artifacts, and fine-grained annotations covering pixel-level segmentation, detailed textual explanations, and artifact category labels. 
Furthermore, we propose \textbf{LEGION} (\textbf{LE}arning to \textbf{G}round and explain for Synthetic \textbf{I}mage detecti\textbf{ON}), a multimodal large language model~(MLLM)-based image forgery analysis framework that integrates artifact detection, segmentation, and explanation.
Building upon this capability, we further explore LEGION as a \textbf{controller}, integrating it into image refinement pipelines to guide the generation of higher-quality and more realistic images. Extensive experiments show that LEGION outperforms existing methods across multiple benchmarks, particularly surpassing the second-best traditional expert on SynthScars by \textbf{3.31\%} in mIoU and \textbf{7.75\%} in F1 score. Moreover, the refined images generated under its guidance exhibit stronger alignment with human preferences. The code, model, and dataset will be released.
\end{abstract}

%% file: sec/1_introduction.tex
\section{Introduction}
\label{sec:introduction}


From GANs~\cite{goodfellow2014generative, karras2019style} to diffusion~\cite{ho2020denoising,nichol2021improved,lu2022dpm,hertz2022prompt,nichol2021glide} and autoregressive models~\cite{tian2024visual,tschannen2024jetformer}, image generation technology has evolved rapidly, producing diverse and detailed synthetic images. While it enhances creativity, simplifies design, and addresses data scarcity, it also poses risks such as privacy violations, copyright disputes, and the spread of misinformation. This duality underscores both its transformative power and the ethical dilemmas it introduces.
Many researchers have focused on the risks of image generation technology, developing methods and benchmarks~\cite{li2024fakebench,ye2024loki} for detecting synthetic images to mitigate societal harm. Yet, a thorough review of synthetic image detection research reveals notable limitations.

\input{figures/intro}

\textbf{(I) Challenges in Synthetic Image Detection Datasets.} 
While datasets represented by OpenForensics~\cite{le2021openforensics} contain a large volume of data, they primarily consist of outdated synthetic images generated using early GAN techniques. These images are of poor quality, riddled with noticeable artifacts, and largely limited to cartoon or anime styles, whose artificial nature is easily discernible. Consequently, models trained on such datasets struggle to effectively detect realistic synthetic images.
The RichHF-18K dataset~\cite{liang2024rich} uses point annotations for artifacts in synthetic images, offering low positional accuracy and poor edge delineation. Meanwhile, tampering-based datasets like SID-Set~\cite{huang2024sida} provide full object contour annotations, but their approach is less generalizable as modern synthetic artifacts typically affect only small object regions.


\textbf{(II) Limitations of Synthetic Image Detection and Artifact Localization Methods.} 
Most traditional methods, such as PAL4VST~\cite{zhang2023perceptual}, rely on low-level structural cues, effectively identifying texture disruptions but struggling with artifacts that require global reasoning, such as violations of physical laws governing lighting and shadows. 
Some works~\cite{huang2024ffaa,li2024forgerygpt,xu2024fakeshield,huang2024sida} have introduced multimodal large language models (MLLMs) to address this problem. However, homogenization of research directions has limited further progress, as these methods predominantly focus more on tampered images with limited exploration of fully synthetic ones, where artifacts are more complex, less constrained by real-world references, and rarely studied from an interpretability perspective.

\textbf{(III) Can Synthetic Image Detection Methods Advance Image Generation?} 
Current synthetic image detection methods mitigate the societal risks posed by image synthesis technologies by identifying and localizing artifacts in synthetic images, thereby positioning the detection technology as a \textbf{\texttt{Defender}}. However, image generation is a double-edged sword; focusing solely on its negative implications does not fully harness the potential of synthetic image detection. From this perspective, we aim to inspire a paradigm shift in the design of synthetic image detection methods—from crafting a defender to developing a \textbf{\texttt{Controller}}. This entails not only detecting and localizing artifacts in synthetic images but also guiding image generators to produce more realistic and natural images. By doing so, we can foster the controlled advancement of image generation technologies.

To address these limitations, we present SynthScars, a challenging dataset for synthetic image detection, excluding outdated, low-quality, and cartoon-style images. It features fine-grained annotations with irregular polygons to precisely outline artifacts, along with detailed classifications and explanations. \textbf{\emph{This dual-layer annotation—spatial and explanatory—elevates the dataset’s value for advancing synthetic image detection research.}}
Moreover, to achieve in-depth interpretability, we propose LEGION, a comprehensive image forgery analysis framework tailored for fully synthetic images. By leveraging the powerful prior knowledge, reasoning and expression ability of MLLMs, it achieves strong generalization across different domains and impressive robustness against various perturbations.
Furthermore, we explore the potential of using forgery explanation as feedback to enhance the generation of higher-quality and more realistic images. Specifically, instead of positioning LEGION as a \textbf{\texttt{Defender}}, we employ it as a \textbf{\texttt{Controller}}, and construct two iterative optimization pipelines via image regeneration and inpainting, respectively. 
For image regeneration, artifact explanations from our model iteratively refine the prompt. For image inpainting, detected artifact masks and corresponding explanations guide region-by-region selective refinement, progressively reducing artifact areas and enhancing image authenticity. The overall comparison with previous methods is illustrated in Figure~\ref{fig:intro}.

The main contributions of this paper are as follows:
\begin{itemize}[leftmargin=0.5cm, itemindent=0cm]
    \item We introduce SynthScars, a challenging dataset for synthetic image detection, featuring high-quality synthetic images with diverse content types, as well as fine-grained pixel-level artifact annotations with detailed textual explanations.
    \item We propose LEGION, a comprehensive image forgery analysis framework for artifact localization, explanation generation, and forgery detection, which effectively aids human experts to detect and understand image forgeries.
    \item Extensive experiments demonstrate that LEGION achieves exceptional performance on four challenging benchmarks. Comparisons with 19 existing methods show that it achieves state-of-the-art performance on the vast majority of metrics, exhibiting strong robustness and generalization ability.
    \item We position LEGION not only as a \textbf{\texttt{Defender}} against ever-evolving generative technologies but also as a \textbf{\texttt{Controller}} that guides higher-quality and more realistic image generation. 
    Qualitative and quantitative experiments on image regeneration and inpainting show the great value of LEGION in providing feedbacks for progressive artifact refinement.
\end{itemize}

%% file: figures/intro.tex
\begin{figure}[!t]
    \centering
    \includegraphics[width=1.02\linewidth]{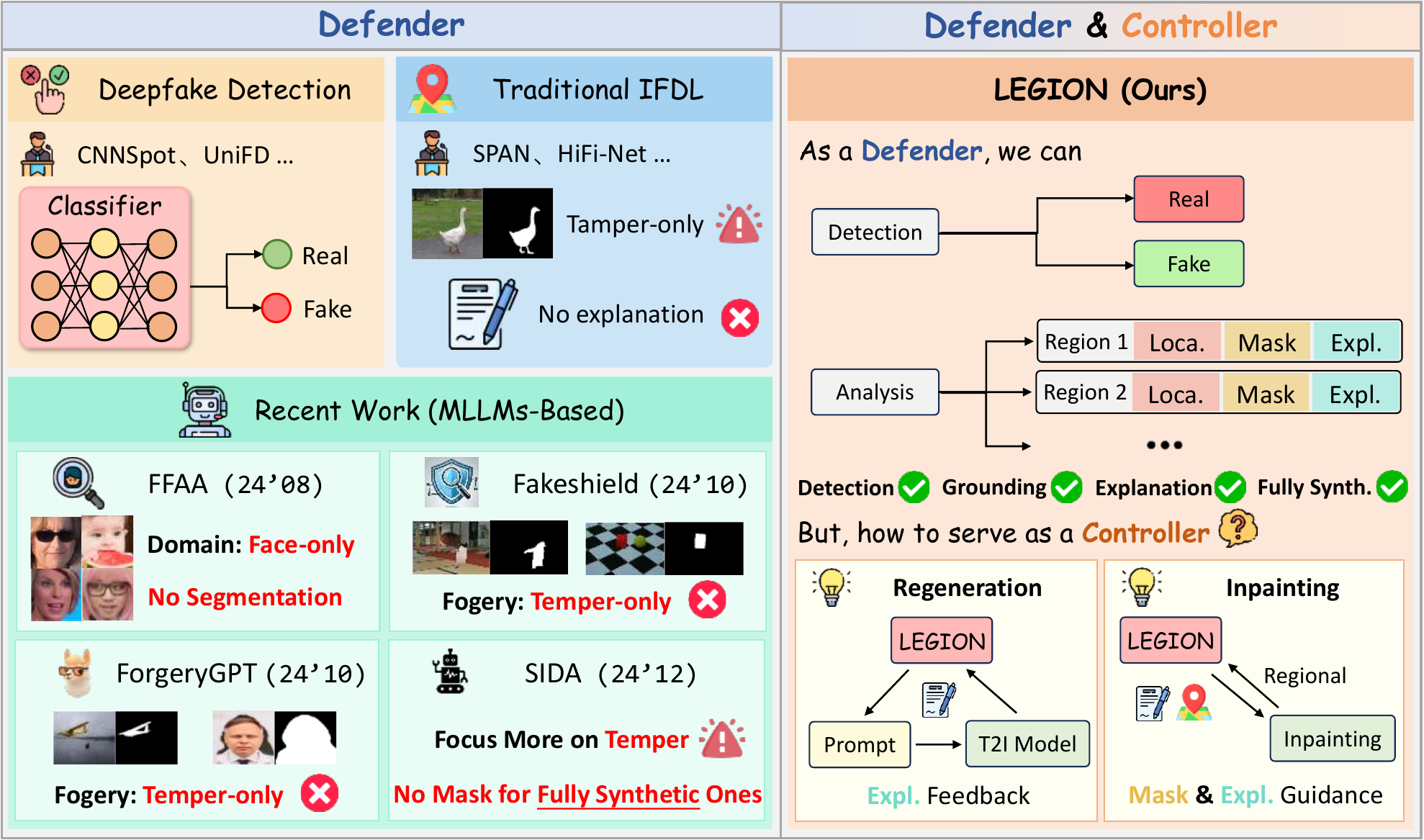}
    \vspace{-6mm}
    \caption{\textbf{Comparison with Existing Image Forgery Detection Methods}. LEGION not only serves as a \textbf{\texttt{Defender}}, enabling multi-task forgery analysis, but also functions as a \textbf{\texttt{Controller}}, facilitating high-quality image generation.}
    \vspace{-6mm}
    \label{fig:intro}
\end{figure}

%% file: sec/2_relatework.tex
\input{figures/dataset_display}
\section{Related Work}
\label{sec:Relate Work}

\subsection{Synthetic Image Detection and Localization}


Traditional detection approaches, based on CNNs and transformers, have treated synthetic image detection as a binary classification task, leveraging spatial or frequency domain features~\cite{chen2024singlesimplepatchneed, yan2024sanity,jeong2022frepgan,tan2024frequency, duan2024test, wang2023dynamic}. However, these methods often struggle with generalization across diverse generators and robustness against various types of perturbations. More importantly, they suffer from a lack of interpretability, with the generation of natural language explanations for specific anomaly causes remaining unexplored.


Recently, many works \cite{zhang2022perceptual,zhang2023perceptual,guo2023hierarchical,guillaro2023trufor} have extended the initial binary classification to a more complicated artifact localization task.  
For example, some studies have utilized gradient \cite{selvaraju2017grad,simonyan2013deep,sundararajan2017axiomatic} or attention maps \cite{huang2024asap} to reveal potential anomalous regions. Alternatively, other scholars have focused on constructing datasets annotated with detailed artifact segmentation masks. However, these methods primarily focus on detecting forgeries from inpainting \cite{zhang2022perceptual} or human manipulation \cite{mazaheri2022detection,wang2022objectformer,huang2022fakelocator}, and overlook the more challenging task of identifying AI-generated traces, which involve inconsistencies across multiple aspects, including image content, structure, style, and other intrinsic features, showing greater flexibility, diversity, and complexity.

\subsection{Multimodal Large Language Models}


Building on the success of large language models (LLMs) \citep{achiam2023gpt, touvron2023llama}, multimodal large language models (MLLMs) \citep{liu2024improved, li2023blip, zhu2023minigpt} extend capabilities by integrating vision and text processing, achieving remarkable performance in comprehensive tasks. 
Some benchmarks, such as FakeBench \cite{li2024fakebench} and LOKI \cite{ye2024loki}, have demonstrated the great potential of MLLMs in synthetic image detection, showing that they can provide more interpretable and context-aware detection results. Moreover, some MLLM-based general visual segmentation models~\cite{lai2024lisa,rasheed2024glamm} have made significant progress, accurately locating objects based on semantic information.

Recent works have explored interpretable synthetic image analysis, providing textual explanations to assist human judgment. 
For instance, FFAA \cite{huang2024ffaa} enhances robustness through its proposed multi-answer intelligent decision system, but is limited to facial data and lacks support for the localization task.
Fakeshield \cite{xu2024fakeshield} carefully designs some modules to generalize across various tampering types, yet lacks analysis for full synthetic images. 
ForgeryGPT \cite{li2024forgerygpt} proposes a customized LLM architecture and a novel framework that captures high-order forensic knowledge correlations of forged images from diverse feature spaces, enabling explainable generation and interactive dialogue. 
Although SIDA explores both tampered and fully synthetic images, it only provides artifact segmentation results for tampered ones.
In summary, most methods focus mainly on tampering analysis and have not explored the more complex task of localizing AI-generated artifacts.

\subsection{Guided Image Refinement}

Existing image generation models support multimodal conditional generation, making text-guided regeneration and artifact-aware inpainting based on segmentation masks and explanations possible.
Early works \cite{brooks2023instructpix2pix,hertz2022prompt} employed text-driven approaches for conditional image generation. Other approaches, such as ControlNet \cite{zhang2023adding}, extended this to support multimodal conditional inputs, such as masks and edge maps, enabling more targeted and controllable refinement processes. 
Recently, methods based on LLMs \cite{yu2023interactive,sun2023dreamsync} have emerged. For instance, Idea2Img \cite{yang2023idea2img} constructs an agent system that leverages GPT-4V's rethinking capability to progressively refine the prompt for text-to-image (T2I) models and iteratively guides image regeneration, improving both image quality and text-image alignment.

Some works have explored region-level image forgery analysis to guide inpainting models in revising anomalous regions. For example, PAL4Inpaint\cite{zhang2022perceptual} and PAL4VST \cite{zhang2023perceptual} inject segmentation masks as conditional input into inpainting models\cite{zhao2021large,suvorov2022resolution} or refiners like SDXL \cite{podell2023sdxl}. However, lacking textual artifact explanations, they often resort to simple methods like object removal, failing to preserve semantics. We extend this idea by integrating our image forgery analysis framework, LEGION, into the pipeline to provide generation models with precise artifact cues.

%% file: figures/dataset_display.tex
\begin{figure*}[!h]
    \centering
    \includegraphics[width=1.02\linewidth]{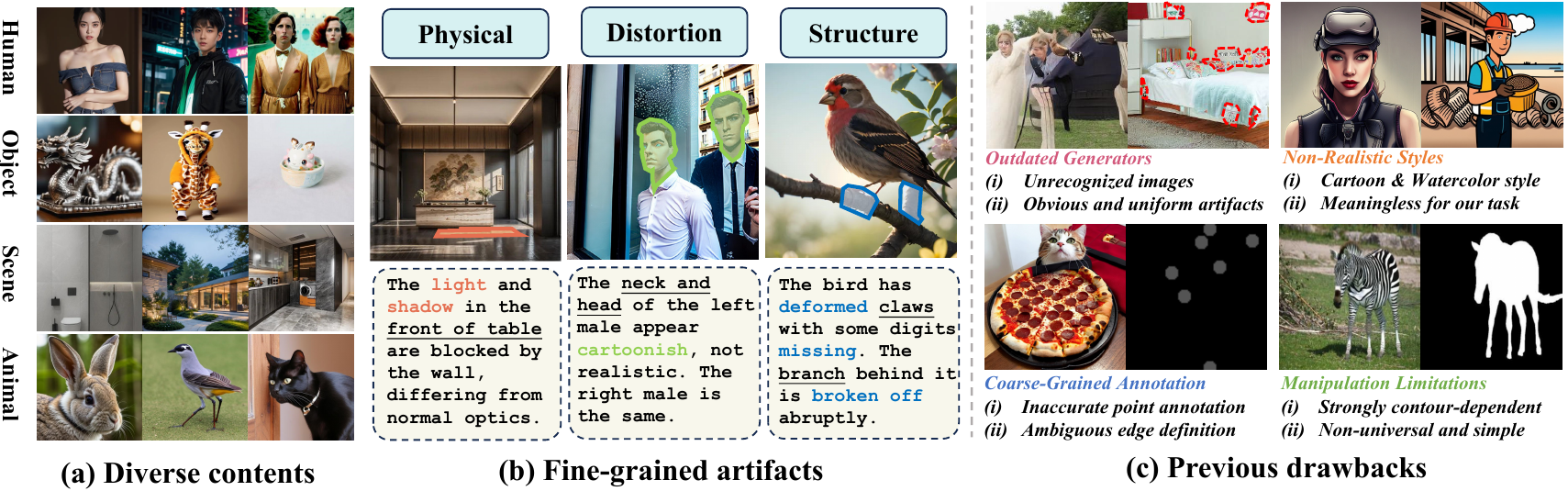}
    \vspace{-2mm}
    \caption{\textbf{SynthScars Datasets.} \textbf{(a)} shows image cases across four diverse content types. \textbf{(b)} presents annotation cases across different fine-grained artifact types. \textbf{(c)} enumerates drawbacks of previous datasets, which SynthScars perfectly addresses.}
    \vspace{-2mm}
    \label{fig:dataset_play}
\end{figure*}

%% file: sec/3_datasets.tex
\section{SynthScars Dataset}
\label{sec:datasets}
\subsection{Motivation}
\label{sec:dataset_motivation}

Recent advancements in generative AI have made it easier to create sophisticated synthetic and tampered content, but existing detection datasets face significant limitations:
\textbf{(i) Outdated Content:} Benchmarks like ProGAN~\citep{gao2019progan} rely on early GANs, producing low-fidelity images easily distinguishable from photorealistic outputs of modern models like Stable Diffusion 3.5\footnote{\url{https://github.com/Stability-AI/sd3.5}} and FLUX\footnote{\url{https://github.com/black-forest-labs/flux}}.
\textbf{(ii) Domain Mismatch:} Some datasets focus on anime-style or synthetic-cartoon images, creating a gap with natural photographic content crucial for real-world applications.
\textbf{(iii) Annotation Issues:} Datasets like RichHF-18K~\citep{liang2024rich} use sparse point annotations, sacrificing spatial precision and introducing ambiguity during training, especially for subtle or complex manipulations.
\textbf{(iv) Contour Dependency:} Recent datasets, such as SID-Set~\citep{huang2024sida}, focus on manipulations with clear-cut boundaries, failing to represent real-world forgeries involving irregular or contextually blended alterations, limiting model generalizability.

\subsection{Dataset Construction}
To address the limitations outlined in Section~\ref{sec:dataset_motivation}, we construct a comprehensive dataset \textbf{SynthScars} through a rigorous pipeline designed to maximize real-world relevance and annotation precision, shown in Figure~\ref{fig:dataset_play}. Specifically, we aggregate and curate samples from multiple public datasets including RichHF-18K~\citep{liang2024rich}, Chameleon~\citep{yan2024sanity}, FFAA~\citep{huang2024ffaa}, and others, following the protocols outlined below. \\
\noindent\textbf{Data Preprocessing and Quality Control.}
To ensure a balanced and high-quality synthetic dataset, we first perform source sampling by clustering latent feature representations from a pretrained ResNet-50~\citep{koonce2021resnet}, followed by uniform sampling from each cluster to mitigate dataset-specific biases while retaining diversity in manipulation types and semantic content. Subsequently, we address limitations (i) and (ii) through a multistage filtering process using Qwen2-VL-72B-Instruct~\citep{wang2024qwen2}, which removes low-quality samples (\emph{e.g.}, blurred or compressed artifacts), non-photorealistic content (\emph{e.g.}, anime-style images), and samples exhibiting conspicuous synthetic patterns.

\noindent\textbf{Dual Fine-grained Annotation.}
To address limitations (iii) and (iv), we adopt irregular polygon masks for annotating artifacts in synthetic images. This fine-grained annotation approach enables precise labeling of artifacts of any shape, size, or location within the image. It also aligns more closely with the characteristics of current synthetic data, where entire objects are rarely artifacts. Compared to point-based annotations, irregular polygon masks offer superior accuracy in localizing artifact regions.
Additionally, \cite{mathys2024synthetic} categorizes artifacts in synthetic images into three types: physics, distortion, and structure. Inspired by this, we incorporate fine-grained artifact categorization into our annotation process, ensuring systematic and precise labeling of different artifact types. This approach enhances the clarity and organization of our dataset while providing a deeper understanding of synthetic artifact characteristics, enabling more targeted analysis and model training.

\subsection{Dataset Statistics}
Compared to existing synthetic image datasets, Table~\ref{tab:datasetcompare} highlights SynthScars' unique combination of pixel-level masks, textual explanations, and artifact-type labels for all samples, setting a new benchmark in synthetic image analysis with 100\% valid annotations.
For detailed dataset statistics, including image categories and artifact annotations, please refer to the Appendix~\ref{app:dataset}.

\input{tables/dataset_compare}



%% file: tables/dataset_compare.tex
\begin{table}[!h]
\centering
\Huge
\resizebox{1.0\linewidth}{!}{
    \begin{tabular}{c|ccccc}
    \toprule[1.5pt]
    \textbf{Dataset}  & \textbf{Pixel-level Mask} & \textbf{Explanation} & \textbf{Artifact Type} & \textbf{Annotator} & \textbf{Valid Sample}  \\
    \midrule
    CNNSpot~\cite{wang2020cnn} & \textcolor{Red}{\XSolidBrush} & \textcolor{Red}{\XSolidBrush} & \textcolor{Red}{\XSolidBrush} & - & 0 \\
    CIFAKE~\cite{bird2024cifake} & \textcolor{Red}{\XSolidBrush} & \textcolor{Red}{\XSolidBrush} & \textcolor{Red}{\XSolidBrush} & - & 0 \\
    UniFD~\cite{ojha2023towards} & \textcolor{Red}{\XSolidBrush} & \textcolor{Red}{\XSolidBrush} & \textcolor{Red}{\XSolidBrush} & - & 0 \\
    GenImage~\cite{zhu2023genimage} & \textcolor{Red}{\XSolidBrush} & \textcolor{Red}{\XSolidBrush} & \textcolor{Red}{\XSolidBrush} & - & 0 \\
    Chamelon~\cite{yan2024sanity} & \textcolor{Red}{\XSolidBrush} & \textcolor{Red}{\XSolidBrush} & \textcolor{Red}{\XSolidBrush} & - & 0 \\
    AI-Face~\cite{lin2024ai} & \textcolor{Red}{\XSolidBrush} & \textcolor{Red}{\XSolidBrush} & \textcolor{Red}{\XSolidBrush} & - & 0 \\
    PAL4VST~\cite{zhang2023perceptual} & \textcolor{Green}{\CheckmarkBold} & \textcolor{Red}{\XSolidBrush} & \textcolor{Red}{\XSolidBrush} & Human & 10168 \\
    RichHF-18K~\cite{liang2024rich} & \textcolor{Red}{\XSolidBrush} & \textcolor{Red}{\XSolidBrush} & \textcolor{Red}{\XSolidBrush} & Human & 11140 \\
    LOKI~\cite{ye2024loki} & \textcolor{Red}{\XSolidBrush} & \textcolor{Green}{\CheckmarkBold} & \textcolor{Red}{\XSolidBrush} & Human & 229 \\
    MMTD-Set~\cite{xu2024fakeshield} & \textcolor{Orange}{\CheckmarkBold\kern-1.2ex\raisebox{1.15ex}{\rotatebox[origin=c]{125}{\textbf{--}}}} & \textcolor{Green}{\CheckmarkBold} & \textcolor{Red}{\XSolidBrush} & GPT-4 & 0 \\
    FF-VQA~\cite{huang2024ffaa} & \textcolor{Red}{\XSolidBrush} & \textcolor{Green}{\CheckmarkBold} & \textcolor{Red}{\XSolidBrush} & GPT-4 & 0 \\
    SID-Set~\cite{huang2024sida} & \textcolor{Orange}{\CheckmarkBold\kern-1.2ex\raisebox{1.15ex}{\rotatebox[origin=c]{125}{\textbf{--}}}} & \textcolor{Green}{\CheckmarkBold} & \textcolor{Red}{\XSolidBrush} & GPT-4 & 0 \\
    \midrule
    \textbf{SynthScars} & \textcolor{Green}{\CheckmarkBold} & \textcolor{Green}{\CheckmarkBold} & \textcolor{Green}{\CheckmarkBold} & \textbf{Human} & \textbf{12236} \\
    \bottomrule[1.5pt]
    \end{tabular}
}
\vspace{-2mm}
\caption{\textbf{Comparison with Existing Image Forgery Datasets.} \underline{The last column} shows the number of samples \textbf{fully synthesized} by common generators, with \textbf{realistic style} and \textbf{valid masks}. \textcolor{Orange}{\CheckmarkBold\kern-1.2ex\raisebox{1.15ex}{\rotatebox[origin=c]{125}{\textbf{--}}}} denotes that only masks of tampered images are provided.}
\vspace{-6mm}
\label{tab:datasetcompare}
\end{table}


%% file: sec/4_method.tex
\input{figures/framework}
\section{Method}
\label{sec:method}
Recent research, such as LISA~\cite{lai2024lisa} and GLaMM~\cite{rasheed2024glamm}, have demonstrated the potential of multimodal large language models~(MLLMs) in performing general image segmentation. Inspired by this work, we introduce the similar idea to the fully synthetic image forgery analysis task to address the limitations discussed in Section~\ref{sec:Relate Work}. To this end, we propose LEGION, a multi-task image forgery analysis framework that supports deepfake detection, artifact localization, and explanation generation. Furthermore, to enhance the quality and realism of current image generation techniques, we build two training-free pipelines based on LEGION for image regeneration and inpainting, respectively.

\subsection{LEGION Architecture}

LEGION consists of four core components: (i) Global Image Encoder, (ii) LLM, (iii) Grounding Image Encoder, and (iv) Pixel Decoder, as shown in Figure~\ref{fig:framework}~(a). Each is described in detail in the following sections. First, to extract global features from the input image, we employ ViT-H/14 CLIP as the \textit{global image encoder}~($\mathcal{E}_{g}$).

\noindent\textbf{Deepfake Detection.} For this task, we follow the conventional approach and formulate it as a binary classification problem, where an image is categorized as either real or fake. We leverage the \texttt{CLS} token from the global feature representation and pass it through a two-layer MLP to predict the detection result. Specifically, given an input image $x_{i}$, it is first encoded into a feature vector $I_{x} = \mathcal{E}_{g}(x_{i}) \in \mathbb{R}^{D_v}$. We then extract the \texttt{CLS} token~(denoted as $\mathrm{CLS}(\cdot)$) from this feature representation and pass it through an MLP classifier~(denoted as $\mathrm{MLP}(\cdot)$) to obtain the probability distribution $y_d$ over the real and fake classes:
\begin{equation}
    y_{d} = \mathrm{MLP}\Bigl(\mathrm{CLS}(I_x)\Bigr).
\end{equation}
\noindent\textbf{Explanation Generation.} To further provide user-friendly natural language explanations, we incorporate a vicuna-based LLM~($\mathcal{L}$) to facilitate vision-language alignment. We design a prompt template: ``\texttt{The <image> provides an overview of the image.}" + $x_{p}$, where $x_{p}$ represents the specialized prompt designed for image forgery analysis. Specifically, we first pass the remaining 256 tokens (excluding the \texttt{CLS} token, denoted as $I^{'}_x$) from the CLIP global image encoder through a \textit{vision-to-language} (V-L) projection layer~($\mathcal{P}_{vl}$), which replaces the \texttt{<image>} token in the prompt template. The processed prompt is then concatenated with the forgery analysis prompt $x_{p}$ shown in Figure~\ref{fig:framework}~(a) to construct the final prompt, guiding the LLM to generate textual explanations $y_e$:
\begin{equation}
    y_{e} = \mathcal{L}\Bigl(x_{p},\mathcal{P}_{vl}(I^{'}_x)\Bigr).
\end{equation}
\noindent\textbf{Artifact Localization.} To obtain pixel-level artifact masks, we apply a pretrained SAM encoder as our \textit{grounding image encoder}~($\mathcal{E}_{l}$) and design our \textit{pixel decoder}~($\mathcal{D}$) with reference to the SAM decoder. The output of the LLM will append a specialized token \texttt{<SEG>} after every description of artifact locations (\emph{e.g.}, the cat's ears), and then a \textit{language-to-prompt}~(L-P) projection layer($\mathcal{P}_{lp}$) is used to transform the text embeddings related to \texttt{<SEG>} token~($v_{seg}$) into the decoder's feature space. Ultimately, $\mathcal{D}$ produces the binary masks~($M$) through the following equation:
\begin{equation}
    M = \mathcal{D}\Bigl(\mathcal{E}_{l}~(x_i),\mathcal{P}_{lp}(v_{seg})\Bigr),\; \text{s.t.}, M_i \in \{0,1\}.
\end{equation}






\noindent\textbf{Training.} We adopt a two-stage independent training strategy. In stage 1, we first train the artifact localization and explanation generation tasks by optimizing segmentation performance using a weighted combination of Binary Cross-Entropy (BCE) and Dice loss, while employing Cross-Entropy (CE) loss to evaluate the discrepancy between the predicted explanation and the ground truth annotations. In stage 2, we focus on enhancing the model’s forgery detection capability using typical CE loss for classification. The loss of each stage can be formulated as:
\begin{equation}
\label{eq:training}
\begin{aligned}
\mathcal{L}_{s1} &= \lambda_{\text{bce}} \mathcal{L}_{\text{BCE}}(M, \hat{M}) + \lambda_{\text{dice}} \mathcal{L}_{\text{Dice}}(M, \hat{M}) \\
&\quad + \lambda_{\text{ce}} \mathcal{L}_{\text{CE}}(y_e, \hat{y}_e), \\
\mathcal{L}_{s2} &= \mathcal{L}_{\text{CE}}(y_d, \hat{y}_d).
\end{aligned}
\end{equation}

\subsection{Image Refinement Pipeline}
Generation and detection technologies mutually reinforce each other, driving their co-evolution. Inspired by~\cite{zhang2023perceptual}, we extend LEGION from a forgery \textbf{\texttt{Defender}} to a generation \textbf{\texttt{Controller}}, enabling artifact-free refinement without additional training data. By providing guidance and feedback to generation models, LEGION facilitates the progressive elimination of potential artifacts, thus enhancing the final output’s quality and realism. To achieve this, we explore two refinement strategies: one leveraging prompt revision for image regeneration, and the other employing inpainting techniques to selectively correct artifact regions.

\input{tables/main_localization}
\noindent\textbf{Regeneration.} We employ an iterative generation approach by combining prompt revision with a text-to-image (T2I) model, as illustrated in Figure~\ref{fig:framework}(b). 
Given an initial image $I_0$ and prompt $P_0$, we feed this prompt into T2I model~(denoted as \textit{Regen}(·)) to regenerate an updated image $I_1$, which may be poorly generated artifact regions. $I_1$ is subsequently analyzed using our LEGION framework, which returns detailed explanations of artifact anomalies~(\emph{e.g.} fingers are deformed). These explanations will be recorded in a memory bank~($\mathcal{M}$). Next, a text refiner~(denoted as \textit{Revise}(·)) revises the initial prompt $P_0$ by rethinking the historical records in the memory bank, enriching the descriptions of regions related to artifacts. 
This results in a revised prompt $P_1$, which is then used to generate the next iteration of the image. This iterative process continues multiple times, progressively refining both the image and prompt. Specifically, for the \textit{t}-th iteration (\textit{t} being a positive integer in the range 0 to N-1), the recurrence formula can be expressed as:
\begin{equation}
    \begin{aligned}
        \mathcal{M}_{t} &= \mathcal{M}_{t-1} + \text{LEGION}(I_t), \\
        P_{t+1} &= \text{Revise}(P_t, \mathcal{M}_t), \\
        I_{t+1} &= \text{Regen}(P_{t+1}).
    \end{aligned}
\end{equation}
\noindent\textbf{Inpainting.} 
We also construct a pipeline to facilitate the inpainting process, as illustrated in Figure~\ref{fig:framework}(c), by using an inpainting model to iteratively remove artifacts and progressively enhance image quality. Compared to regeneration, this approach better preserves non-artifact regions since the image is not entirely regenerated but selectively refined  only for anomalous areas. Specifically, given an input image $I_0$, we first perform image forgery analysis using our LEGION framework, and organize the feedbacks into a region-wise triplet set~($\mathcal{A}$), where each region is represented as $(L_{i},M_{i},E_{i})$, with $L$, $M$, $E$ denoting the location, mask, and explanation, respectively. Each identified artifact region is then refined using the inpainting model~(denoted as \textit{Inpaint}(·)) by feeding corresponding mask and textual explanation into the model. This is a region-by-region process until all regions are done, finally yielding the refined image. By iterating the above process multiple times, we can obtain higher-quality and more realistic images. For the image $I_t$ at the \textit{t}-th iteration, this process can be formulated as:
\small
\begin{equation}
    \begin{aligned}
    \mathcal{A}_t &= \text{LEGION}(I_t) = \left\{ (L_i, M_i, E_i) \mid i = 1,2,\dots,k \right\} \\ 
    I_{t+1} &= I_t + \sum_{i=1}^{k} M_i \cdot \bigl( \text{Inpaint}(I_t, M_i, E_i) - I_t \bigr).
    \end{aligned}
\end{equation}
\normalsize

%% file: figures/framework.tex
\begin{figure*}[!ht]
    \centering
    \includegraphics[width=1.02\linewidth]{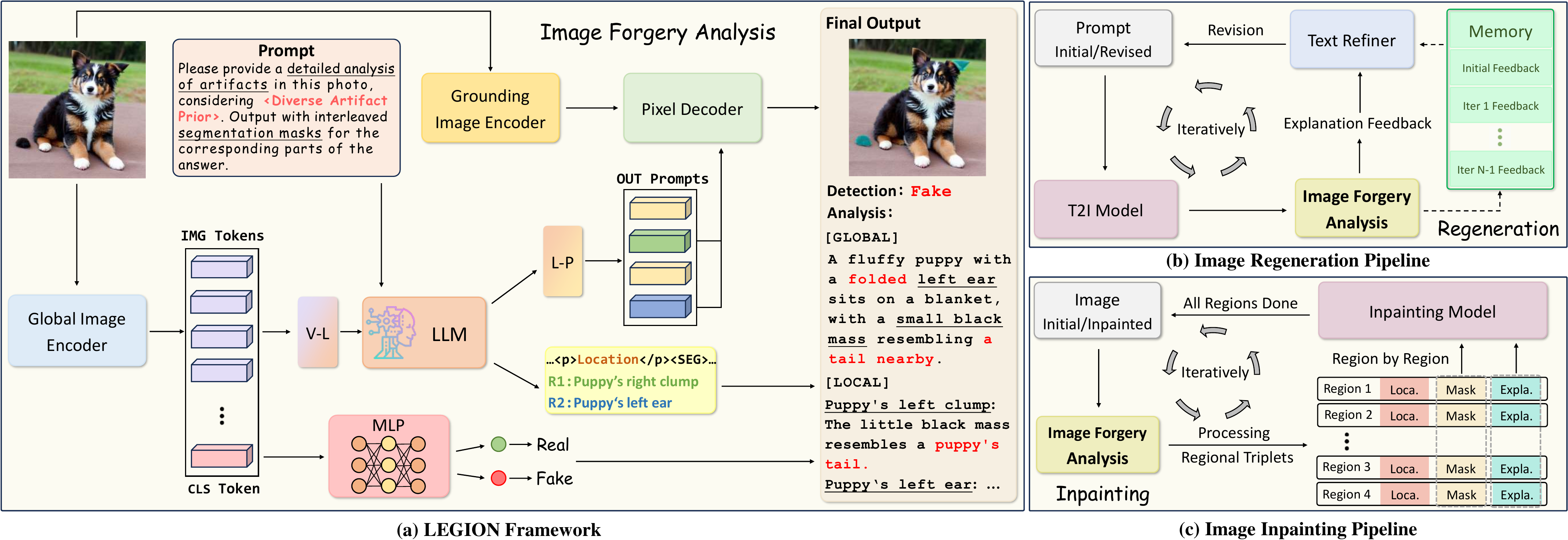}
    \vspace{-4mm}
    \caption{\textbf{Architecture Overview.} \textbf{(a)} Our proposed framework for image forgery analysis, LEGION. \textbf{(b)} and \textbf{(c)} shows two pipelines for image generation. T2I in (b) is short for \textit{text-to-image}, and Loca. and Expla. in (c) denotes \textit{Location} and \textit{Explanation}, respectively.}
    \vspace{-4mm}
    \label{fig:framework}
\end{figure*}

%% file: tables/main_localization.tex
\renewcommand{\multirowsetup}{\centering}
\definecolor{mygray}{gray}{.92}
\definecolor{ForestGreen}{RGB}{34,139,34}
\newcommand{\fg}[1]{\mathbf{\mathcolor{ForestGreen}{#1}}}
\definecolor{Forestred}{RGB}{220,50,50}
\newcommand{\fr}[1]{\mathbf{\mathcolor{Forestred}{#1}}}
\begin{table*}[!ht]
    \centering
    \belowrulesep=-0.25pt
    \aboverulesep=-0.25pt
    \setlength{\tabcolsep}{4pt}
    \renewcommand{\arraystretch}{1.22}
    \footnotesize
	\centering
    \begin{tabular}{c| >{\centering\arraybackslash}p{1.0cm} | >{\centering\arraybackslash}p{0.7cm} >{\centering\arraybackslash}p{0.7cm} >{\centering\arraybackslash}p{0.7cm} >{\centering\arraybackslash}p{0.7cm} >{\centering\arraybackslash}p{0.7cm} >{\centering\arraybackslash}p{0.7cm} >{\centering\arraybackslash}p{0.7cm} >{\centering\arraybackslash}p{0.7cm} | >{\centering\arraybackslash}p{0.7cm} >{\centering\arraybackslash}p{0.7cm} | >{\centering\arraybackslash}p{0.7cm} >{\centering\arraybackslash}p{0.7cm} }
        \toprule[1.2pt]
        \multirow{3}{*}{\textbf{Method}} & \multirow{3}{*}{\textbf{Source}} & \multicolumn{8}{c|}{\textbf{SynthScars}} & \multicolumn{2}{c|}{\multirow{2}{*}{\textbf{LOKI}}} & \multicolumn{2}{c}{\multirow{2}{*}{{\textbf{RichHF-18K}}}} \\
        & & \multicolumn{2}{c}{\textbf{Object}} & \multicolumn{2}{c}{\textbf{Animal}} & \multicolumn{2}{c}{\textbf{Human}} & \multicolumn{2}{c|}{\textbf{Scene}} & & & & \\
        \cmidrule(l{3pt}r{3pt}){3-4} \cmidrule(l{3pt}r{3pt}){5-6} \cmidrule(l{3pt}r{3pt}){7-8} \cmidrule(l{3pt}r{3pt}){9-10} \cmidrule(l{3pt}r{3pt}){11-12} \cmidrule(l{3pt}r{3pt}){13-14}
        & & \textbf{mIoU} & \textbf{F1} & \textbf{mIoU} & \textbf{F1} & \textbf{mIoU} & \textbf{F1} & \textbf{mIoU} & \textbf{F1} & \textbf{mIoU} & \textbf{F1} & \textbf{mIoU} & \textbf{F1} \\
        \hline
        
        HiFi-Net$^\text{\faStarO}$~\cite{guo2023hierarchical} & \texttt{CVPR23} & 43.74 & 0.45 & 45.28 & 0.03 & 46.21 & 0.84 & 45.90 & 0.04 & 39.60 & 2.41 & 44.96 & 0.39 \\

         TruFor$^\text{\faStarO}$~\cite{guillaro2023trufor} & \texttt{CVPR23} & 46.99 & 14.82 & 48.45 & 17.57 & 49.02 & 15.43 & 48.93 & 12.64 & 46.55 & \underline{16.70} & 48.41 & \underline{18.03} \\

         PAL4VST$^\text{\faStarO*}$~\cite{zhang2023perceptual} & \texttt{ICCV23} & \underline{50.46} & 19.25 & \underline{52.55} & \underline{21.61} & \underline{59.18} & \underline{35.70} & \underline{52.55} & \underline{19.14} & \underline{47.34} & 11.58 & \underline{49.88} & 14.78 \\
         \hline
         \textcolor{gray}{Ferret$^{\bm\square}$~\cite{you2023ferret}} & \textcolor{gray}{\texttt{ICLR24}} & \textcolor{gray}{30.10} & \textcolor{gray}{20.81} & \textcolor{gray}{27.17} & \textcolor{gray}{15.78} & \textcolor{gray}{25.54} & \textcolor{gray}{13.87} & \textcolor{gray}{30.64} & \textcolor{gray}{13.79} & \textcolor{gray}{24.50} & \textcolor{gray}{18.88} & \textcolor{gray}{26.52} & \textcolor{gray}{16.22} \\

          \textcolor{gray}{Griffon$^{\bm\square}$~\cite{zhan2024griffon}} & \textcolor{gray}{\texttt{ECCV24}} & \textcolor{gray}{38.54} & \textcolor{gray}{23.40} & \textcolor{gray}{27.76} & \textcolor{gray}{18.58} & \textcolor{gray}{23.04} & \textcolor{gray}{14.81} & \textcolor{gray}{35.83} & \textcolor{gray}{14.47} & \textcolor{gray}{21.96} & \textcolor{gray}{20.41} & \textcolor{gray}{28.13} & \textcolor{gray}{18.19} \\
         
         LISA-v1-7B$^\text{\faStarO*}$~\cite{lai2024lisa} & \texttt{CVPR24} & 35.49 & \underline{23.70} & 32.44 & 18.77 & 34.11 & 17.50 & 37.56 & 18.31 & 31.10 & 9.29 & 35.90 & \textbf{21.94} \\
         \hline
         InternVL2-8B$^{\bm\square}$~\cite{chen2024internvl} & \texttt{CVPR24} & 41.08 & 13.36 & 41.22 & 7.83 & 41.21 & 3.91 & 41.68 & 7.55 & 42.03 & 10.06 & 39.90 & 9.58 \\
         
         \textcolor{gray}{Qwen2-VL-72B$^{\bm\square}$~\cite{wang2024qwen2}} & \textcolor{gray}{-} & \textcolor{gray}{33.89} & \textcolor{gray}{23.25} & \textcolor{gray}{32.46} & \textcolor{gray}{21.98} & \textcolor{gray}{26.92} & \textcolor{gray}{14.75} & \textcolor{gray}{39.00} & \textcolor{gray}{18.17} & \textcolor{gray}{26.62} & \textcolor{gray}{20.99} & \textcolor{gray}{27.58} & \textcolor{gray}{19.02} \\
         \hline
         \rowcolor{green!8}
        \textbf{LEGION$^\text{\faStarO}$~(Ours)} & - & \textbf{54.62} & \textbf{29.90} & \textbf{54.52} & \textbf{27.43} & \textbf{60.82} & \textbf{39.44} & \textbf{53.67} & \textbf{24.51} & \textbf{48.66} & \textbf{16.71} & \textbf{50.07} & 17.41 \\

        \bottomrule[1.2pt]
	\end{tabular}
     \vspace{-2mm}
    \caption{\textbf{Performance Comparison of Artifact Localization on SynthScars and Two Benchmarks in Unseen Domains.} $*$ denotes methods fine-tuned on SynthScars, while others use pre-trained weights due to unavailable training code. $\text{\faStarO}$ and $\bm\square$ represent the models that output segmentation masks and bounding boxes, respectively. \textcolor{gray}{Grayed approaches} predict most of the image as artifacts and are only for reference, not included in the comparison.}
    \label{tab:mainlocalization}
     \vspace{-2mm}
\end{table*}

%% file: sec/5_experiments.tex
\section{Experiments}
\label{sec:experiments}

\subsection{Experimental Setup}
\noindent\textbf{Implementation Details}
We adopt the pretrained weights of GLaMM due to its strong capability in the \textit{grounded conversation generation} task it proposed. In training stage 1, we fine-tune GLaMM using LoRA on 8 NVIDIA A100 GPUs with $\alpha$ = 8 and a batch size of 2 per device. The initial learning rate is set to 1e-4, and the weight hyperparameters $\lambda_{ce}$, $\lambda_{dice}$, and $\lambda_{bce}$ in Eq~\ref{eq:training}, which balance the three types of loss, are set to 1.0, 0.2, and 0.4, respectively. In training stage 2, we directly train the MLP on ProGAN using 8 NVIDIA A100 GPUs, with an initial learning rate of 1e-3 and a batch size of 64 per device.

\noindent\textbf{Evaluation Metrics.}
For the artifact localization task, we evaluate the segmentation performance by reporting the mean Intersection over Union (mIoU) of foreground and background regions, as well as the overall F1 scores, following~\cite{zhang2023perceptual,guo2023hierarchical,guillaro2023trufor}. Additionally, to assess the alignment between the generated textual explanations and ground truth, we employ two metrics: ROUGE-L, which is based on n-gram overlap and sequence matching, and Cosine Similarity Score (CSS), which computes the cosine similarity between text embeddings obtained from a pre-trained language model, following FakeShield~\cite{xu2024fakeshield}~(more details in Appendix~\ref{app:explanation}). In the image refinement section, we assess the quality and realism of regenerated and inpainted images using the Human Preference Score (HPS) proposed in~\cite{wu2023human}.

\subsection{Localization Evaluation}
To evaluate artifact localization performance, we use training split of SynthScars as training data and test on its test split for in-domain assessment. We further examine models' generalization performance to unseen domains using LOKI~\cite{ye2024loki} and RichHF-18K~\cite{liang2024rich}, which have been filtered to retain only images with realistic styles.
We compare LEGION with state-of-the-arts (SOTAs) for fully synthetic artifact localization. Baselines include traditional expert models such as HiFi-Net~\cite{guo2023hierarchical}, TruFor~\cite{guillaro2023trufor}, and PAL4VST~\cite{zhang2023perceptual}, as well as VLMs for object-grounding, including Ferret~\cite{you2023ferret}, Griffon~\cite{zhan2024griffon}, and LISA~\cite{lai2024lisa}. Moreover, we benchmark general-purpose MLLMs that have shown great performance across various tasks, including InternVL2~\cite{chen2024internvl}, Qwen2-VL~\cite{wang2024qwen2} and DeepSeek-VL2~\cite{wu2024deepseekvl2mixtureofexpertsvisionlanguagemodels}.


Results in Table~\ref{tab:mainlocalization} demonstrate that LEGION achieves SOTA performance across all three evaluation datasets, despite its F1 score on RichHF-18K being slightly lower than that of LISA-v1-7B and TruFor. Compared with traditional experts, LEGION outperforms the strongest expert model, PAL4VST, by \textbf{10.65} points in F1 score for \textit{Object} category on SynthScars, and also consistently surpassing it on the other two datasets.
For object-grounding VLMs and general-purpose MLLMs, we observe that these models struggle with artifact localization due to the lack of pre-training and prior knowledge specific to this task. This leads to two extreme behaviors: some models fail to identify foreground regions altogether (\emph{e.g.}, DeepSeek-VL2), while others overestimate artifacts, treating most of the image as a huge artifact (\emph{e.g.}, Ferret, Griffon, and Qwen2-VL), resulting in low mIoU but artificially high F1 scores due to extreme foreground recall. InternVL2 and LISA fine-tuned on SynthScars exhibit these extremes less severely, yet LEGION still outperforms both across the majority of metrics. The visualization comparison of various methods is presented in Figure~\ref{fig:good_cases}.

\noindent\textbf{Robustness Study.} We also systematically compare the localization performance between LEGION and PAL4VST under three types of perturbations. The results can be found in Table~\ref{tab:robustlocalization} of Appendix~\ref{app:more_experiments}, indicating the robustness of our model under strong interference conditions—a critical capability unattainable by conventional approaches.


\input{figures/good_cases}
\subsection{Explanation Performance}
\input{tables/text_evaluation}
To better assess the model interpretability and capability to generate natural language explanations for artifacts, we conduct a comparative analysis of the latest released open-source (\emph{e.g.}, DeepSeek-VL2) and closed-source models (\emph{e.g.}, December,2024 updated GPT-4o) with varying parameters.
In our evaluation, we test on SynthScars and LOKI which contain detailed artifact explanations, measuring ROUGE-L for surface-level structural alignment and CSS for semantic equivalence to jointly assess both lexical coherence and contextual fidelity. Since SynthScars has a fixed response format and LEGION trained on it demonstrates structural advantages, we convert the outputs of others to the same format for fair comparison. The experimental results are presented in Table~\ref{tab:textevaluation}, indicating the superior performance of LEGION across both datasets.


\input{tables/hps_comparision}
\subsection{Detection Performance}
\input{tables/main_detection}
Following previous works~\cite{zhang2019detecting,chai2020makes,ojha2023towards,tan2024frequency,tan2024rethinking}, we train LEGION on ProGAN~\cite{gao2019progan} and evaluate its cross-generator generalization on the UniversalFakeDetect benchmark~\cite{ojha2023towards}. As shown in Table~\ref{tab:maindetection}, LEGION achieves the highest accuracy on GANs, CRN, and IMLE, secures competitive second-place accuracy on SITD, and maintains comparable detection performance in other generators.

\subsection{Image Refinement Cases}
To evaluate LEGION's effectiveness in guiding generation models toward higher-quality, more realistic images, we randomly sample 200 images from the SynthScars test set. Following~\cite{chen2024mj}, we compute the average HPS scores for images before and after refinement (for both regeneration and inpainting) and measure the average growth rate of each refinement process. The results are summarized in Table \ref{tab:hpscomparison}.

\input{figures/regen_case}
\subsubsection{Prompt Revision and Image Regeneration}
We consider the inherent gap between the original image generator and the advanced refiner Stable Diffusion 3.5, which may independently lead to significant quality improvements unrelated to LEGION's guidance. To control external influences, we first generate a concise one-sentence description of the original image using GPT-4o, then use Stable Diffusion 3.5 to regenerate the image as the refinement starting point. Finally, we perform two rounds of iterative regeneration to produce the final result.
Figure~\ref{fig:regen_case} showcases two image regeneration cases.  
\textbf{Case 1}: LEGION detects a cartoonish style in the original image and refines the prompt with constraints like \textit{``natural lighting"} and \textit{``realistic style"}. After one refinement round, the image becomes significantly more realistic.  
\textbf{Case 2}: The woman’s left pinky finger is deformed in the initial image. Guided by LEGION, subsequent prompts add hand-specific details, enabling the model to refine the region. Two rounds of optimization are needed to correct the structure and achieve a natural form.

\subsubsection{Image Inpainting under Feedback Guidance}
Leveraging the powerful inpainting capability of the SD-XL model\footnote{\url{https://huggingface.co/diffusers/stable-diffusion-xl-1.0-inpainting-0.1}}, we explore region-level image restoration guided by LEGION’s feedback, including artifact masks and anomaly explanations. For each image, we perform 3 iterations of the inpainting pipeline.
Figure~\ref{fig:inpaint_case} illustrates the iterative refinement process for correcting reflections in a synthetic image. In the original image, the left reflection on the water mismatches the wall's color, while the right reflection contains an unrealistic window shape, violating physical laws. Through multiple iterations, LEGION progressively identifies entire reflection region, highlighting color and shape discrepancies to guide the inpainting process. By the 3-rd iteration, the artifacted region is successfully refined, achieving high-quality restoration.
Notably, such physical artifacts requiring global reasoning are often challenging for previous localization models to detect. This underscores LEGION’s strong capability in addressing artifact localization in fully synthetic images.

\input{figures/inpaint_case}

%% file: figures/good_cases.tex
\begin{figure}[!t]
    \centering
    \includegraphics[width=1.02\linewidth]{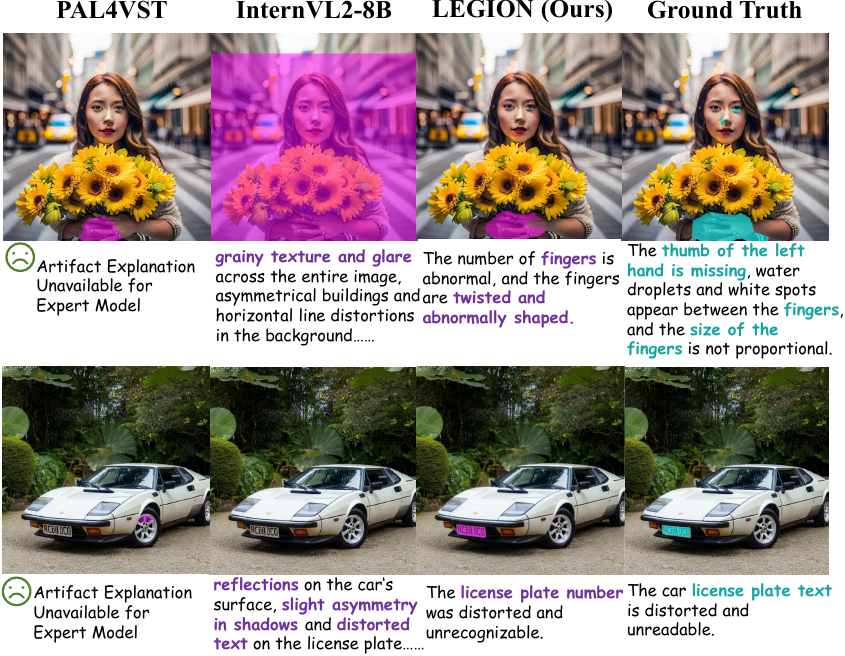}
    \caption{\textbf{Comparison of artifact segmentation and explanations across different methods}: PAL4VST, InternVL2-8B, and our proposed LEGION, alongside the ground truth.}
    \vspace{-4mm}
    \label{fig:good_cases}
\end{figure}

%% file: tables/text_evaluation.tex

\renewcommand{\multirowsetup}{\centering}
\definecolor{ForestGreen}{RGB}{34,139,34}
\begin{table}[!h]
    \centering
    \belowrulesep=-0.25pt
    \aboverulesep=-0.25pt
    \setlength{\tabcolsep}{1.0pt}
    \renewcommand{\arraystretch}{1.3}
    \footnotesize
	\centering
    \resizebox{1.0\linewidth}{!}{
    \begin{tabular}{c| >{\centering\arraybackslash}p{1cm} | >{\centering\arraybackslash}p{1cm} | >{\centering\arraybackslash}p{1.5cm} >{\centering\arraybackslash}p{1.4cm} | >{\centering\arraybackslash}p{1.5cm} >{\centering\arraybackslash}p{1.4cm} }
        \toprule[1.2pt]
        \multirow{2}{*}{\textbf{Method}} & \multirow{2}{*}{\textbf{Date}} & \multirow{2}{*}{\textbf{Params}} & \multicolumn{2}{c|}{\textbf{SynthScars}} & \multicolumn{2}{c}{\textbf{LOKI}}\\
        \cmidrule(l{3pt}r{3pt}){4-5} \cmidrule(l{3pt}r{3pt}){6-7}
        & & & \textbf{ROUGE-L$\uparrow$} & \textbf{CSS$\uparrow$} & \textbf{ROUGE-L$\uparrow$} & \textbf{CSS$\uparrow$} \\
        \hline

        Qwen2-VL~\cite{wang2024qwen2} & 24.09 & 72B & 25.84 & 58.15 & 11.80 & 37.64 \\
        LLaVA-v1.6~\cite{liu2024llavanext} & 24.01 & 7B & \underline{29.61} & \underline{61.75} & \underline{16.07} & \underline{41.07} \\
        InternVL2~\cite{chen2024internvl} & 24.07 & 8B & 25.93 & 56.89 & 10.10 & 39.62 \\
        Deepseek-VL2~\cite{wu2024deepseekvl2mixtureofexpertsvisionlanguagemodels} & 24.12 & 27B & 25.50 & 47.77 & 6.70 & 28.76 \\

        GPT-4o~\cite{hurst2024gpt} & 24.12 & - & 22.43 & 53.55 & 9.61 & 38.98 \\

        \rowcolor{green!8}
        \textbf{LEGION~(Ours)} & \textbf{25.03} & \textbf{8B} & \textbf{39.50} & \textbf{72.60} & \textbf{18.55} & \textbf{45.96} \\
        
        \bottomrule[1.2pt]
	\end{tabular}}
     \vspace{-2mm}
    \caption{\textbf{Comparison of Multimodal Models in Artifact Explanation Generation.} Metrics are normalized to the range of 0–100 for better visualization and comparison.}
    \label{tab:textevaluation}
     \vspace{-2mm}
\end{table}

%% file: tables/hps_comparision.tex
\renewcommand{\multirowsetup}{\centering}
\definecolor{ForestGreen}{RGB}{34,139,34}
\begin{table}[!h]
    \centering
    \belowrulesep=-0.25pt
    \aboverulesep=-0.25pt
    \setlength{\tabcolsep}{2.5pt}
    \renewcommand{\arraystretch}{1.15}
    \footnotesize
	\centering
    \begin{tabular}{c| >{\centering\arraybackslash}p{2cm} | >{\centering\arraybackslash}p{2cm} }
        \toprule[1.2pt]
         \textbf{HPS} & \textbf{Regeneration} & \textbf{Inpainting} \\
        \hline

        Pre-refined Score~(Avg.) & 31.24 & 29.57 \\
        Post-refined Score~(Avg.) & 33.36 & 30.20 \\
        \hline

        Growth Rate $\uparrow$ & \textbf{6.98\%} & \textbf{2.14\%} \\
        
        \bottomrule[1.2pt]
	\end{tabular}
     \vspace{-2mm}
    \caption{\textbf{HPS Comparison Before and After Refinement in Regeneration and Inpainting.} Scores are normalized to the range of 0–100 for better visualization and comparison.}
    \label{tab:hpscomparison}
\end{table}

%% file: tables/main_detection.tex
\renewcommand{\multirowsetup}{\centering}
\definecolor{mygray}{gray}{.92}

\renewcommand{\multirowsetup}{\centering}
\definecolor{mygray}{gray}{.92}
\begin{table}[!t]
    \centering
    \belowrulesep=-0.25pt
    \aboverulesep=-0.25pt
    \setlength{\tabcolsep}{2.0pt}
    \renewcommand{\arraystretch}{1.2}
    \footnotesize
    \centering
    \resizebox{1.02\linewidth}{!}{
    \begin{tabular}{c| c | c | c c | c c | c }
        \toprule[1.2pt]
        \multirow{2}{*}{\textbf{Method}} & \multirow{2}{*}{\textbf{GANs}} & \multirow{2}{*}{\textbf{Deepfakes}} & \multicolumn{2}{c|}{\textbf{Perceptual Loss}} & \multicolumn{2}{c|}{\textbf{Low Level Vision}} & \multirow{2}{*}{\textbf{Diffusion}} \\
        & & & \textbf{CRN} & \textbf{IMLE} & \textbf{SITD} & \textbf{SAN} & \\
        \hline
        
        Co-occurence~\cite{nataraj2019detecting}  & 75.17 & 59.14 & 73.06 & \underline{87.21} & 68.98 & 60.42 & 85.53 \\

        Freq-spec~\cite{zhang2019detecting} & \underline{75.28} & 45.18 & 53.61 & 50.98 & 47.46 & 57.12 & 69.00 \\

        CNNSpot~\cite{wang2020cnn} & 85.29 & 53.47 & \underline{86.31} & 86.26 & 66.67 & 48.69 & 58.63 \\

        Patchfor~\cite{chai2020makes} & 69.97 & 75.54 & 72.33 & 55.30 & 75.14 & 75.28 & 72.54    \\

        UniFD~\cite{ojha2023towards} & 95.25 & 66.60 & 59.50 & 72.00 & 63.00 & 57.50 & 82.02   \\

        LDGard~\cite{tan2023learning} & 89.17 & 58.00 & 50.74 & 50.78 & 62.50 & 50.00 & \underline{89.79} \\

        FreqNet~\cite{tan2024frequency} & 94.23 & \textbf{97.40} & 71.92 & 67.35 & \textbf{88.92} & 59.04 & 83.34 \\

        NPR~\cite{tan2024rethinking} & 94.16 & \underline{76.89} & 50.00 & 50.00 & 66.94 & \textbf{98.63} & \textbf{94.54} \\


        \rowcolor{green!8}
        \textbf{LEGION~(Ours)} & \textbf{97.01} & 63.37 & \textbf{90.78} & \textbf{98.93} & \underline{79.44} & 57.76 & 83.10 \\ 

        \bottomrule[1.2pt]
	\end{tabular}}
     \vspace{-2mm}
    \caption{\textbf{Comparison of Synthetic Image Detection on UniversalFakeDetect Benchmark.} All methods are trained on ProGAN.}
    \label{tab:maindetection}
     \vspace{-3mm}
\end{table}

%% file: figures/regen_case.tex
\begin{figure}[!t]
    \centering
    \includegraphics[width=1.0\linewidth]{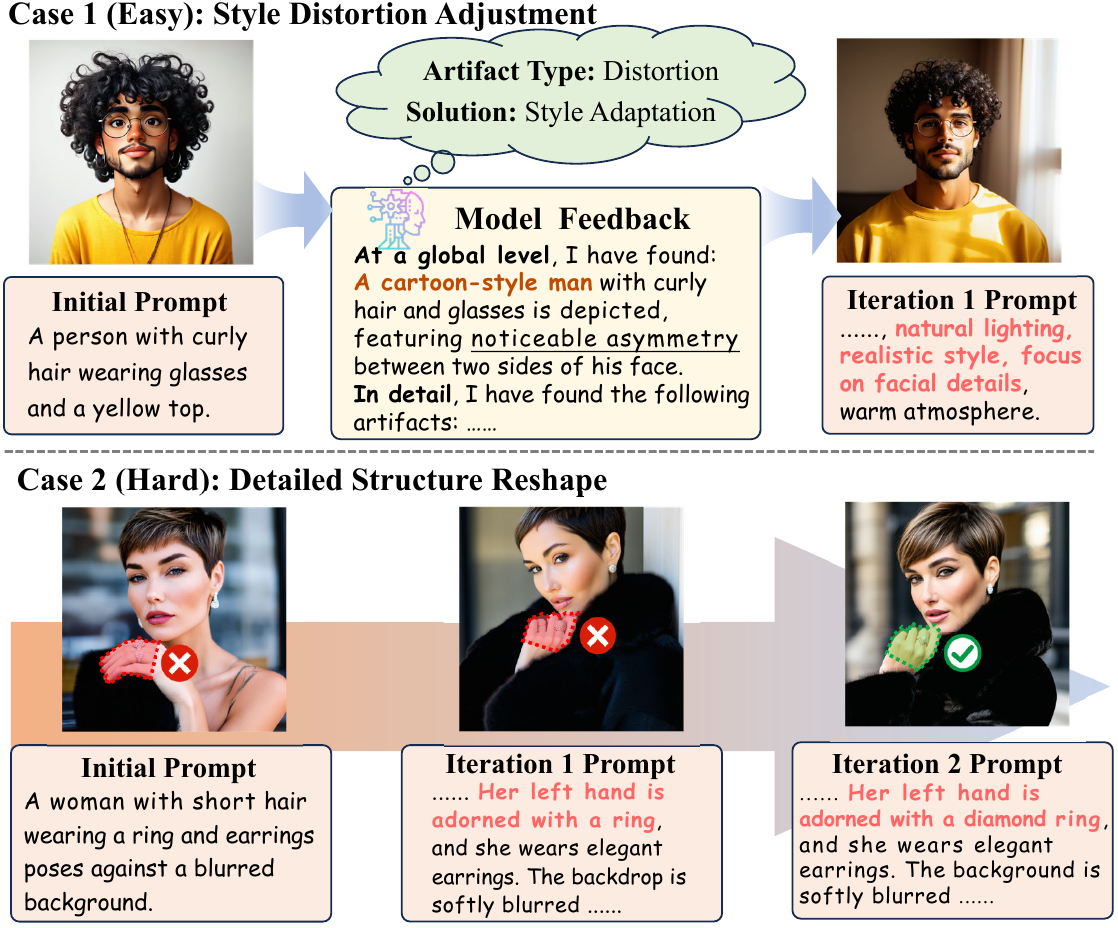}
    \vspace{-4mm}
    \caption{\textbf{Case studies of Image Regeneration}. \textbf{(Top)} Style Distortion Adjustment, \textbf{(Bottom)} Detailed Structure Reshape.}
    \vspace{-4mm}
    \label{fig:regen_case}
\end{figure}


%% file: figures/inpaint_case.tex
\begin{figure}[!t]
    \centering
    \includegraphics[width=1.0\linewidth]{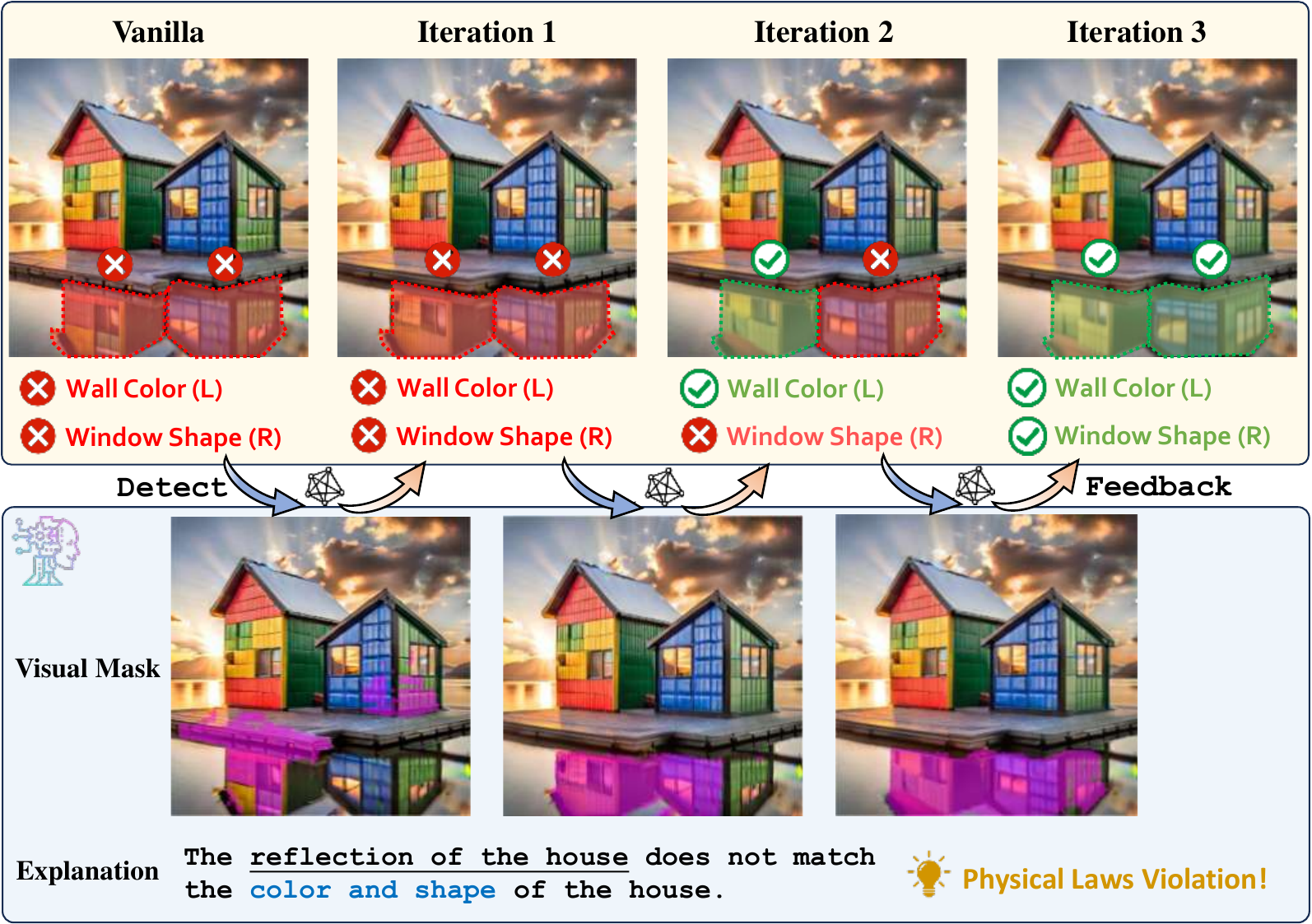}
    \vspace{-4mm}
    \caption{\textbf{Case Study of Image Inpainting}.}
    \vspace{-4mm}
    \label{fig:inpaint_case}
\end{figure}


%% file: sec/6_conclusion.tex
\section{Conclusion}
\label{sec:conclusion}


In this work, we explore the challenging task of fully synthetic image forgery analysis by introducing SynthScars dataset, which features diverse and high-difficulty artifact instances that are not constrained by local contours and require global understanding. To enable more fine-grained artifact analysis, we propose LEGION, a MLLM-based framework that enhances interpretability compared to traditional methods. Experimental results show that it achieves superior performance and strong robustness across multiple benchmarks and evaluation metrics. It also demonstrates strong potential as a controller for guided image generation and inpainting from both qualitative and quantitative perspectives.
While our proposed dataset and method address some key limitations in image forgery analysis, the inherent diversity and flexibility of fully synthetic artifacts leave ample room for further improvements in the future. We also hope more researchers to explore this domain and contribute to the responsible and ethical use of generative AI.

%% file: sec/appendix.tex


\section{Appendix}
\appendix
\textbf{Contents of the Appendices:}

Section~\ref{app:dataset}. Detailed Construction Process and Statistics of the SynthScars Dataset.

Section~\ref{app:more_experiments}. More Experimental Details for Image Forgery Analysis Task.

Section~\ref{app:rubustness_study}. Robustness Comparison Between LEGION and Expert Model under Various Perturbations.

Section~\ref{app:case_comparison}. Additional Visual Examples of LEGION.

Section~\ref{app:limitations}. Limitations and Failure Case Analysis.

\section{SynthScars Dataset}\label{app:dataset}
\subsection{Artifact Definition}
Inspired by \cite{mathys2024synthetic}, we categorize artifacts in synthetic images into three types: physics, distortion, and structure. To eliminate subjective differences among annotators and to clarify and standardize the criteria for artifact classification during the annotation process, we established a guideline that explicitly defines the nature and scope of various artifacts, as shown in Table~\ref{tab:artifact_definition}.

\subsection{Annotation Details}
We recruited 12 experienced annotators with high-education backgrounds and provided them with dedicated training before the annotation task. They were required to strictly follow the guideline for data annotation and discard samples where artifacts were entirely imperceptible to human eyes. 
The annotation process for 12,236 samples in SynthScars took a total of 240 hours and underwent multiple rounds of quality inspection.

\subsection{Data Curation}
To obtain high-quality, deceptive, and challenging synthetic images, we carry out a multistage filtering process using Qwen2-VL-72B-Instruct~\citep{wang2024qwen2}, which removes low-quality samples (\emph{e.g.}, blurred or compressed artifacts), non-photorealistic content (\emph{e.g.}, cartoonish or watercolor-style images), and samples exhibiting conspicuous synthetic patterns. Specifically, we designed a prompt, as shown in the Table~\ref{tab:curation_table}, for the model to sequentially inspect each data sample against the given criteria. 
Only samples that meet all the standards are retained.

\input{tables/content_category}
\input{tables/artifacts_type}
\subsection{Dataset Statistics}
As shown in Table~\ref{tab:contentstatistics}, SynthScars includes 12,236 fully synthesized images across diverse real-world scenarios, with 11,236 training and 1,000 test samples categorized into human, object, animal, and scene. The dataset features 26,566 artifact instances (Table~\ref{tab:artifactstatistics}), annotated with irregular polygon masks and classified into three types: physics-related (6\%), distortion (5\%), and structural anomalies (89\%).

\section{Experimental Details}\label{app:more_experiments}
\subsection{Prompt Design}
When designing the prompt, in order to fully unleash the LLM's broad reasoning ability, we incorporated prior knowledge of different artifacts (denoted as \texttt{<Diverse Artifact Prior>}). Specifically, it consists of common cases from the three types of artifacts we defined, guiding the model to examine the image from the corresponding perspectives. To provide a concrete example, we define it as follows:
\begin{takeaways}
      ~\textbf{Physics artifacts} (\emph{e.g., optical display issues, violations of physical laws, and spatial/perspective errors}), \textbf{Structure artifacts} (\emph{e.g., deformed objects, asymmetry, or distorted text}), and \textbf{Distortion artifacts} (\emph{e.g., color/texture distortion, noise/blur, artistic style errors, and material misrepresentation})
\end{takeaways}

\input{tables/artifact_definition}
\input{tables/curation}

\subsection{Explanation Evaluation}\label{app:explanation}
Following Fakeshield~\cite{xu2024fakeshield}, we use paraphrase-MiniLM-L6-v2\footnote{\url{https://huggingface.co/sentence-transformers/paraphrase-MiniLM-L6-v2}} from HuggingFace as our text embedding model to transform the outputs into semantic feature space.

\input{tables/robust_localization}

\section{Robustness Study}\label{app:rubustness_study}
We compare the artifact localization performance between LEGION and PAL4VST (the strongest expert model from Table~\ref{tab:mainlocalization}) on SynthScars under three types of distortion. 
Table~\ref{tab:robustlocalization} reveal that Gaussian noise induces the most severe performance degradation, followed by Gaussian blur, while JPEG compression exhibits the least negative effects. Notably, as intensity increases, LEGION remains stable, while PAL4VST degrades sharply, highlighting our model's superior robustness under strong interference—an unattainable ability for traditional expert models.

\section{More Visual Examples}\label{app:case_comparison}
In addition to comparing LEGION's predictions with other methods, including multi-modal large language models and expert models, this section provides an extended visualization of artifact segmentation masks and their corresponding explanations. 
As shown in Figure~\ref{app_fig:good_cases}, LEGION excels in predicting artifacts on highly realistic synthetic images, achieving both positional and contour accuracy in segmentation. The accompanying explanations are insightful, highlighting not only the location of the artifact but also offering a plausible rationale for its artificial nature. These results highlight LEGION's ability to deliver precise artifact detection alongside interpretable insights, enhancing the transparency and trustworthiness of synthetic image generation.

\section{Limitations and Analysis}\label{app:limitations}
While our model demonstrates promising results in detecting and segmenting artifacts on AI-generated images, there remain areas for improvement. A qualitative analysis of failure cases reveals two primary challenges.
First, in scenarios with high scene complexity and a multitude of elements, our model sometimes tends to miss subtle artifact regions. As illustrated in Figure~\ref{app_fig:bad_cases}, the predicted masks may incompletely cover the areas affected by anomalies, particularly when the artifacts are intertwined with intricate background details.
Second, the model struggles with detecting very subtle artifacts that occupy a small image area, especially in human portraits. These artifacts, often manifesting as minor distortions or unnatural textures, can be difficult to perceive even for the human eye.
We argue that the model is being overwhelmed by the sheer volume of information, leading to a prioritization of more prominent anomalies at the expense of smaller, less conspicuous ones.
\input{figures/appendix_good_cases}
\input{figures/bad_cases}

%% file: tables/content_category.tex
\begin{table}[!h]
\centering
\resizebox{0.98\linewidth}{!}{
    \begin{tabular}{c|cccc|c}
    \toprule
    \textbf{Image Content}      & \textbf{Human} & \textbf{Object} & \textbf{Animal} & \textbf{Scene} & \textbf{Total} \\
    \midrule
    \textbf{Train} & 6253 & 1940 & 1183 & 1860 & 11236 \\
    \textbf{Test} & 587 & 162 & 134 & 117 & 1000 \\
    \midrule
    \textbf{Total} & 6840 & 2102 & 1317 & 1977 & 12236 \\
    \bottomrule
    \end{tabular}
}
\vspace{-2mm}
\caption{\textbf{Statistics on Image Content.} SynthScars encompasses a diverse range of real-world scenarios, including 12,236 fully synthesized images from different generators.}
\vspace{-2mm}
\label{tab:contentstatistics}
\end{table}

%% file: tables/artifacts_type.tex
\begin{table}[!h]
\centering
\resizebox{0.98\linewidth}{!}{
    \begin{tabular}{c|ccc|c}
    \toprule
    \textbf{Artifact Type}      & \textbf{Physics} & \textbf{Distortion} & \textbf{Structure} & \textbf{Total} \\
    \midrule
    \textbf{Train} & 1431 & 1249 & 21233 & 23913 \\
    \textbf{Test} & 111 & 136 & 2406 & 2653 \\
    \midrule
    \textbf{Total} & 1542 & 1385 & 23639 & 26566 \\
    \bottomrule
    \end{tabular}
}
\vspace{-2mm}
\caption{\textbf{Statistics on Artifact Types.} SynthScars classifies artifacts into three fine-grained anomaly types, and contains a total of 26,566 artifact instances.}
\label{tab:artifactstatistics}
\vspace{-4mm}
\end{table}

%% file: tables/artifact_definition.tex
\begin{table*}[htb]\centering
    \begin{minipage}{\textwidth}\vspace{0mm}    \centering
    \begin{tcolorbox} 
    \centering
    \small
    \begin{tabular}{p{0.95\textwidth} c}
    \VarSty{ {\bf \normalsize Artifact Definition} } &\\
    

    \begin{enumerate}[leftmargin=*]
        \setlength{\itemsep}{1pt}  
        \setlength{\parskip}{1pt}  
        \item \textbf{Physics}
        \begin{enumerate}
            \item \textbf{Optical Display}: These artifacts arise from inconsistencies in the propagation and reflection of light, violating fundamental optical principles. They can occur across different objects and scenes, leading to unrealistic visual effects. Common cases include incorrect reflections, shadows, and light source positioning errors, causing synthetic images to deviate from real-world optical phenomena.
            \item \textbf{Physical Law Violations}: These artifacts result from the failure to adhere to fundamental physical laws during image synthesis. They typically manifest as illogical scenes, such as water flowing upward or objects floating in mid-air, which contradict natural laws.
            \item \textbf{Space and Perspective}: These artifacts stem from inaccuracies in object proportions and spatial relationships during image generation, leading to inconsistencies with real-world perspective rules. Examples include incorrect depth perception, mismatched object sizes, or spatial distortions that prevent accurate perspective alignment.
        \end{enumerate}
    
        \item \textbf{Structure}
        \begin{enumerate}
            \item \textbf{Deformed Objects}: These artifacts arise when the shape or structure of objects is distorted due to errors in the generative model. Contributing factors include geometric inconsistencies, texture mapping errors, and rendering issues.
            \item \textbf{Asymmetrical Objects}: These artifacts occur when an object exhibits unnatural asymmetry, deviating from expected structural balance.
            \item \textbf{Incomplete/Redundant Structures}: These artifacts appear as missing or excessive structural components, leading to unrealistic representations of objects.
            \item \textbf{Illogical Structures}: These artifacts involve the generation of unrecognizable or non-existent objects, as well as the appearance of elements that should not logically exist within the given context.
            \item \textbf{Text Distortion and Illegibility}: These artifacts include warped, irregular, or unrecognizable text, affecting the readability and coherence of textual content within the generated image.
        \end{enumerate}
    
        \item \textbf{Distortion}
        \begin{enumerate}
            \item \textbf{Color and Texture}: These artifacts result from errors in color rendering or color space conversion, leading to unnatural hues, inappropriate saturation, or other inconsistencies in color perception.
            \item \textbf{Noise and Blurring}: These artifacts are associated with image noise reduction and clarity enhancement processes. They may arise when algorithms fail to effectively remove noise or introduce excessive blurring, causing local details to appear distorted or unnatural.
            \item \textbf{Artistic Style}: These artifacts occur when synthetic images exhibit unintended stylization, such as cartoonish or painterly appearances that deviate from realistic textures. Such distortions are often caused by errors in style transfer or texture generation algorithms.
        \end{enumerate}
    \end{enumerate}

    \end{tabular}
    \end{tcolorbox}
    \vspace{-3mm}
    \caption{\textbf{Artifact Definition}. We clearly define three types of artifacts and require annotators to strictly follow this guideline for annotation.}
    \label{tab:artifact_definition}
    \end{minipage}
    \vspace{-3mm}
\end{table*}

%% file: tables/curation.tex
\begin{table*}[htb]\centering
    \begin{minipage}{\textwidth}\vspace{0mm}    \centering
    \begin{tcolorbox} 
    \centering
    \small
    \begin{tabular}{p{0.95\textwidth} c}
    

    \VarSty{ {\bf \normalsize  System Prompt} } &\\
    You are a helpful assistant. Analyze the given images based on the following three criteria and assign one label to each image. You only need to return the label for each image without providing any additional explanations: 
    \\ \textbf{\small \textit{Evaluation Criteria}}
    \begin{enumerate}[leftmargin=*]
        \setlength{\itemsep}{1pt}  
        \setlength{\parskip}{1pt}  
        \item \textbf{Clarity}
        \begin{enumerate}
            \item The image should be well-lit, sharp, and visually clear without blurriness, noise, or distortion.
            \item The image must not show obvious signs of artificial manipulation, such as pixelated edges or unnatural distortions.
        \end{enumerate}
        \item \textbf{Safety}
        \begin{enumerate}
            \item The image must not contain violence, blood, gore, explicit sexual content, hate symbols, discriminatory elements, or any harmful or inappropriate material.
            \item Any content that could evoke strong negative emotions or discomfort should be classified as unsafe.
        \end{enumerate}
        \item \textbf{Realism}
        \begin{enumerate}
            \item The image should look realistic and have a photo-like appearance.
            \item It must not be cartoonish, animated, or heavily stylized in an artistic manner.
        \end{enumerate}
    \end{enumerate}
    \textbf{\small \textit{Labeling Task}} \\
    Assign one of the following labels to each image:
    \begin{enumerate}[leftmargin=*]
        \item \textbf{Acceptable}: If the image meets all three criteria;
        \item \textbf{Rejected[Clarity]}: If the image is unclear, blurry, or distorted;
        \item \textbf{Rejected[Safety]}: If the image contains unsafe or inappropriate content;
        \item \textbf{Rejected[Realism]}: If the image is stylized, animated, or lacks realism.
    \end{enumerate}
    \hrulefill & \\
    \VarSty{ {\bf \normalsize User Prompt} } &\\
    Please strictly follow the instructions to label the input image: \{\textit{image}\} \\
    \end{tabular}
    \end{tcolorbox}
    \vspace{-2mm}
    \caption{\textbf{Curation Prompt}. Only samples that meet all the standards are retained.}
    \label{tab:curation_table}
    \end{minipage}
\end{table*}

%% file: tables/robust_localization.tex
\renewcommand{\multirowsetup}{\centering}
\definecolor{ForestGreen}{RGB}{34,139,34}
\begin{table}[!h]
    \centering
    \belowrulesep=-0.25pt
    \aboverulesep=-0.25pt
    \setlength{\tabcolsep}{2pt}
    \renewcommand{\arraystretch}{1.2}
    \footnotesize
    \centering
    \resizebox{1.0\linewidth}{!}{
    \begin{tabular}{c| >{\centering\arraybackslash}p{1.65cm}  >{\centering\arraybackslash}p{1.65cm} | >{\centering\arraybackslash}p{1.65cm} >{\centering\arraybackslash}p{1.65cm} }
        \toprule[1.2pt]
        \multirow{2}{*}{\textbf{Distortion}} & \multicolumn{2}{c|}{\textbf{PAL4VST}} & \multicolumn{2}{c}{\textbf{LEGION~(Ours)}}\\
        \cmidrule(l{3pt}r{3pt}){2-3} \cmidrule(l{3pt}r{3pt}){4-5}
        & \textbf{mIoU} & \textbf{F1} & \textbf{mIoU} & \textbf{F1} \\
        \hline

        No Distortion & 56.10 & 29.21 & 59.41 & 36.96 \\
        \hline

        JPEG Comp.~(QF = 50) & 55.95 & 28.85 & 57.78 & 33.97 \\
        JPEG Comp.~(QF = 35) & 55.55 & 27.60 & 58.04 & 34.08 \\
        JPEG Comp.~(QF = 20) & 55.01 \textbf{\textcolor{ForestGreen}{\scriptsize (-1.9\%)} } & 26.36 \textbf{\textcolor{red!80}{\scriptsize (-9.8\%)} } & 57.91 \textbf{\textcolor{red!80}{\scriptsize (-2.5\%)} } & 34.28 \textbf{\textcolor{ForestGreen}{\scriptsize (-7.3\%)} } \\
        \hline

        Gaussian Noise~($\sigma$ = 0.1) & 56.01 & 28.96 & 57.31 & 33.00 \\
        Gaussian Noise~($\sigma$ = 0.2) & 54.42 & 25.16 & 56.77 & 32.52 \\
        Gaussian Noise~($\sigma$ = 0.3) & 52.91 \textbf{\textcolor{red!80}{\scriptsize (-5.7\%)} } & 21.11 \textbf{\textcolor{red!80}{\scriptsize (-27.7\%)} } & 56.49 \textbf{\textcolor{ForestGreen}{\scriptsize (-4.9\%)} } & 32.12 \textbf{\textcolor{ForestGreen}{\scriptsize (-13.1\%)} } \\
        \hline

        Gaussian Blur~(Ksize = 5) & 55.62 & 27.76 & 57.75 & 33.78 \\
        Gaussian Blur~(Ksize = 9) & 54.58 & 25.23 & 57.27 & 32.63 \\
        Gaussian Blur~(Ksize = 15) & 53.24 \textbf{\textcolor{red!80}{\scriptsize (-5.1\%)}} & 22.30 \textbf{\textcolor{red!80}{\scriptsize (-23.7\%)}} & 57.50 \textbf{\textcolor{ForestGreen}{\scriptsize (-3.2\%)}} & 33.52 \textbf{\textcolor{ForestGreen}{\scriptsize (-9.3\%)}}\\

        \bottomrule[1.2pt]
	\end{tabular}}
    \caption{\textbf{Robustness Comparison Under Different Perturbations.} LEGION significantly outperforms the strongest existing expert model under severe JPEG compression~(denoted as JPEG Comp.), Gaussian noise, and Gaussian blur~(Ksize represents kernel size). Values in parentheses indicate degradation ratios, with the \underline{more robust method} highlighted in \textbf{\textcolor{ForestGreen}{green}}, otherwise in \textbf{\textcolor{red!80}{red}}.}
    \label{tab:robustlocalization}
\end{table}

%% file: figures/appendix_good_cases.tex
\begin{figure*}[!h]
    \centering
    \includegraphics[width=1.02\linewidth]{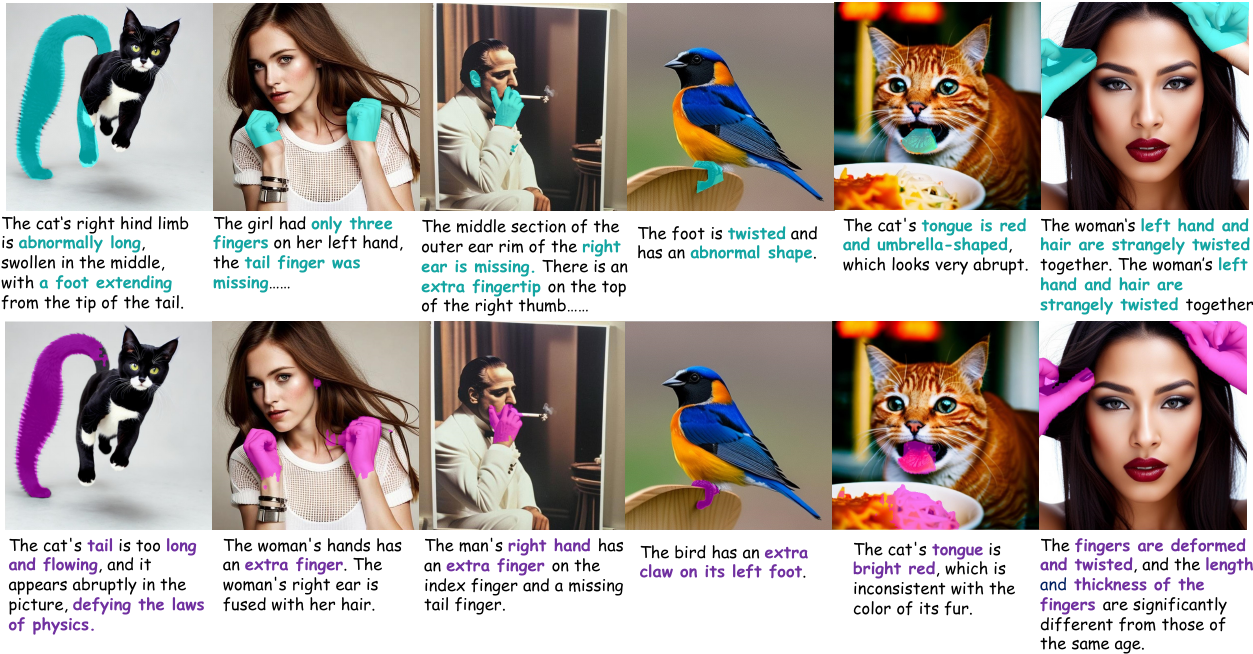}
    \caption{More Visualization of Artifact Segmentation Masks and Corresponding Explanations for Identified Artifacts. The figure illustrates a qualitative comparison between the ground truth (\textbf{Top row}) and the corresponding predictions obtained from our proposed model (\textbf{Bottom row}).}
    \label{app_fig:good_cases}
\end{figure*}

%% file: figures/bad_cases.tex
\begin{figure*}[!h]
    \centering
    \includegraphics[width=1.02\linewidth]{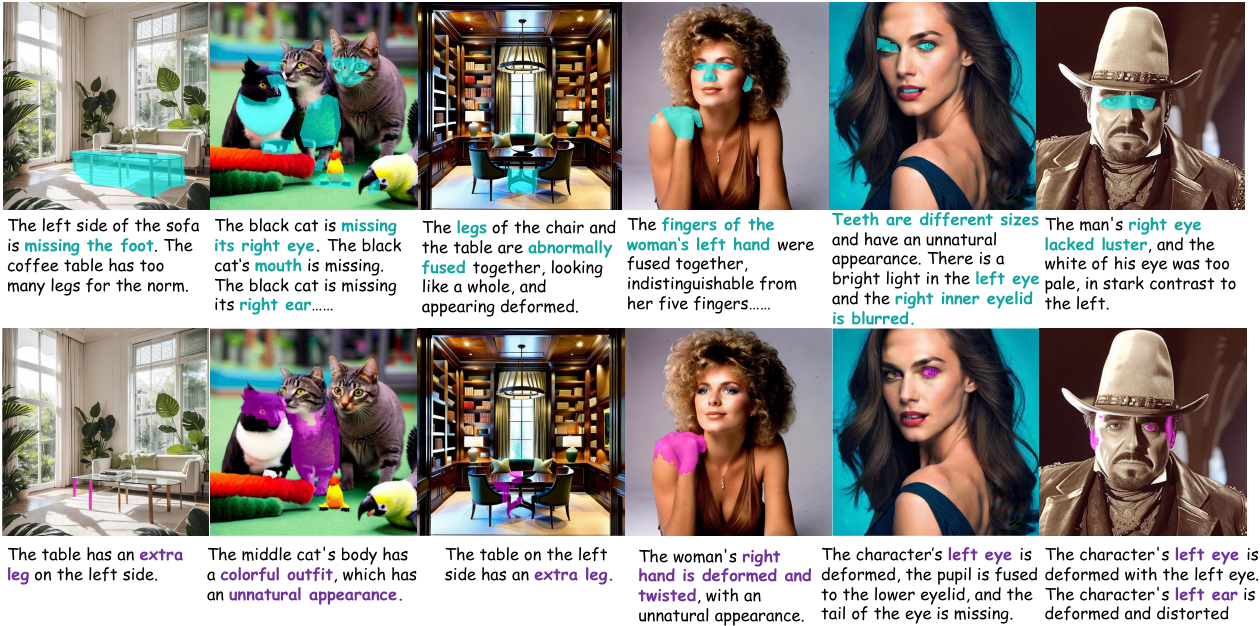}
    \caption{\textbf{Examples of failures in complex scenes and intricate small artifacts.} Each case includes artifacts segmentation mask and corresponding explanations. The first row depicts the ground truth, while the second row shows the corresponding predictions generated by our model.}
    \label{app_fig:bad_cases}
\end{figure*}